\documentclass[10pt,final,journal,twocolumn]{IEEEtran}

\usepackage{booktabs}
\usepackage{amsmath}
\usepackage{mathrsfs}
\usepackage{subfigure}
\usepackage{hyperref}
\usepackage{graphicx}
\usepackage{latexsym}
\usepackage{amssymb,amsbsy,verbatim,array}
\usepackage{epsfig}
\usepackage{epstopdf}
\usepackage{cite}
\usepackage{color}
\usepackage[ruled,linesnumbered]{algorithm2e}
\usepackage{makecell}
\usepackage{extarrows}
\usepackage{multicol}
\usepackage{bm}

\makeatletter
\newcommand{\algorithmfootnote}[2][\footnotesize]{%
  \let\old@algocf@finish\@algocf@finish
  \def\@algocf@finish{\old@algocf@finish
    \leavevmode\rlap{\begin{minipage}{\linewidth}
    #1#2
    \end{minipage}}%
  }%
}
\makeatother

\makeatletter
\newcommand{\tabincell}[2]{
}
\makeatother

\begin{document}

\title{Learning Based Signal Detection for MIMO Systems with Unknown Noise Statistics}

\author{Ke He, Le He, Lisheng Fan, Yansha Deng, George K. Karagiannidis, \emph{Fellow, IEEE}, and  Arumugam Nallanathan, \emph{Fellow, IEEE}
\thanks{K. He, L. He and L. Fan are all with the School of Computer Science and Cyber Engineering, Guangzhou University, China (e-mail: heke2018@e.gzhu.edu.cn, hele20141841@163.com, lsfan@gzhu.edu.cn).}
\thanks{Y.  Deng  is  with  the  Department  of  Informatics, King's  College  London, London WC2R 2LS, UK (e-mail: yansha.deng@kcl.ac.uk).}
\thanks{G. K. Karagiannidis is with the Wireless Communications Systems Group (WCSG), Aristotle University of Thessaloniki, Thessaloniki 54 124, Greece (e-mail: geokarag@auth.gr).}
\thanks{A. Nallanathan is with the School of Electronic Engineering and Computer Science, Queen Mary University of London, London, U.K (e-mail: a.nallanathan@qmul.ac.uk).}
}

\pagestyle{headings}

\maketitle

\thispagestyle{empty}

\begin{abstract}
This paper aims to devise a generalized maximum likelihood (ML) estimator to robustly detect signals with unknown noise statistics in multiple-input multiple-output (MIMO) systems. In practice, there is little or even no statistical knowledge on the system noise, which in many cases is non-Gaussian, impulsive and not analyzable. Existing detection methods have mainly focused on specific noise models, which are not robust enough with unknown noise statistics. To tackle this issue, we propose a novel ML detection framework to effectively recover the desired signal. Our framework is a fully probabilistic one that can efficiently approximate the unknown noise distribution through a normalizing flow. Importantly, this framework is driven by an unsupervised learning approach, where only the noise samples are required. To reduce the computational complexity, we further present a low-complexity version of the framework, by utilizing an initial estimation to reduce the search space. Simulation results show that our framework outperforms other existing algorithms in terms of bit error rate (BER) in non-analytical noise environments, while it can reach the ML performance bound in analytical noise environments. The code of this paper is available at \url{https://github.com/skypitcher/manfe}.
\end{abstract}


\section{Introduction}
Consider the linear inverse problem encountered in signal processing, where the aim is to recover a signal vector $\bm{x} \in \mathbb{C}^{N \times 1}$ given the noisy observation $\bm{y} \in \mathbb{C}^{M \times 1}$, and the channel response matrix $\bm{H} \in \mathbb{C}^{M \times N}$. Formally, the observation vector can be expressed as
\begin{align}
\label{eq:linearmodel}
\bm{y} = \bm{H}\bm{x}+\bm{w},
\end{align}
where $\bm{w} \in \mathbb{C}^{M \times 1}$ is an additive measurement noise, that is independent and identically distributed (i.i.d) with an unknown distribution $p_{\bm{w}}(\bm{w})$. From a Bayesian perspective, the optimal solution to the above problem is the maximum a posteriori (MAP) estimation
\begin{align}
\hat{\bm{x}}_{MAP} =& \arg\max_{\bm{x} \in \mathcal{X}} p(\bm{x}|\bm{y}), \\
=& \arg\max_{\bm{x} \in \mathcal{X}} p(\bm{y}|\bm{x}) p(\bm{x}),
\end{align}
where $\mathcal{X}$ denotes the set of all possible signal vectors. When there is no prior knowledge on the transmitted symbols, the MAP estimate is equivalent to the maximum likelihood estimation (MLE), which can be expressed as
\begin{align}
\hat{\bm{x}}_{MAP} = \hat{\bm{x}}_{MLE} =& \arg\max_{\bm{x} \in \mathcal{X}} p(\bm{y}|\bm{x}), \\
=& \arg\max_{\bm{x} \in \mathcal{X}} p_{\bm{w}}(\bm{y}-\bm{H}\bm{x}).
\label{eq:map}
\end{align}
In most of the existing works in the literature, the noise $\bm{w}$ is assumed to be additive white Gaussian noise (AWGN), whose probability density function (PDF) is analytical and the associated likelihood of each possible signal vector is tractable. In this case, the MLE in (\ref{eq:map}) becomes
\begin{align}\label{eq:mld}
\hat{\bm{x}}_{\text{E-MLE}} = \arg\min_{\bm{x} \in \mathcal{X}} \| \bm{y} - \bm{H}\bm{x} \|^2,
\end{align}
which aims to minimize the Euclidean distance, referred to as E-MLE. However, in practical communication scenarios, we may have little or even no statistical knowledge on the noise. In particular, the noise may present some impulsive characteristics and may be not analyzable. For example, the noise distribution becomes unknown and mainly impulsive for scenarios like long-wave, underwater communications, and multiple access systems \cite{ilow1998analytic, beaulieu2010uwb, Chen2015Suboptimum, Chen2015Novel}. In these cases, the performance of E-MLE will deteriorate severely \cite{PappiKK13}. In contrast to the Gaussian case, the exact PDF of impulsive noise is usually unknown and not analytical \cite{Chen2015Suboptimum, Chen2015Novel, Fan2012On, Samorodnitsky1996Stable}, which means  that the exact likelihood $p(\bm{y}|\bm{x})$ is computationally intractable.

\subsection{Related Research}
In general, there are two major approaches to solve the problem of  signal detection in MIMO systems:  \textit{model-driven } and \textit{data-driven}. Next, we briefly present both of them.
\subsubsection{Model-Driven Methods}
Model-driven approaches have been extensively studied in the literature for  MIMO signal detection, by assuming that the noise is Gaussian. Among them, the approximate message passing (AMP) algorithm is an attractive method, which assumes Gaussian noise and well-posed channel matrix \cite{donoho2009message}. The AMP  can detect the desired signal by iteratively predicting and minimizing the mean squared error (MSE) with a state evolution process \cite{donoho2009message, Rangan2010Generalized}. Combined with deep learning methods, AMP is unfolded into a number of neural layers to improve the performance with ill-posed channel matrix \cite{he2020model, LiuLHYG19}. Moreover, an efficient iterative MIMO detector has been proposed in \cite{LaiLAMC15} to leverage the channel-aware local search (CA-LS) technology to significantly improve the signal detection under Gaussian noise environment. When the noise has an arbitrary density function, a generalized AMP (GAMP) algorithm can be designed for the generalized linear mixing model \cite{Rangan2010Generalized}, where the sum-product version of the GAMP algorithm can be treated as a hybrid of the iterative soft-threshold algorithm (ISTA) and alternating direction method of multipliers (ADMM) \cite{rangan2016fixed}. The Gaussian GAMP (G-GAMP) is equivalent to the AMP, and the GAMP with MMSE denoiser can be rigorously characterized with a scalar state evolution whose fixed points, when unique, are Bayes-optimal \cite{Adel2013State, liu2019capacity}. Since the GAMP algorithm extends the AMP algorithm to adapt to arbitrary noise whose PDF is analytical, it still requires numerical or approximate methods to compute the marginal posterior, when the noise statistics is unknown \cite{Rangan2010Generalized}. In addition, for some specific non-Gaussian noises, A. Mathur \textit{et al.} have investigated the system performance and found the ML detectors in \cite{MathurB14, saxena2017ber, MathurBP15}, which is critical for the development of the model-driven methods.

When there is no prior knowledge on the noise statistics, researchers proposed other model-driven approaches to approximate the unknown noise distribution $p_{\bm{w}}(\bm{w})$ and compute the approximate likelihood $p(\bm{y}|\bm{x})$ accordingly. However, this requires a huge amount of computation or sampling loops. As a result, these methods can not be applied efficiently in practical scenarios. For example, the expectation-maximization (EM) algorithm will be extremely slow in this case, since the dimension of data can be very high and the data set can be very large \cite{Sammaknejad2019A}. Moreover, in order to select an appropriate approximate model, the EM algorithm requires to have some knowledge on the noise statistics, otherwise it may performs worst\cite{Vila2013}. Besides, for the variational inference based approximations in \cite{Tzikas2008, Yu2017Modulation, Zhang2017One}, the noise is assumed to depend on a hidden variable $\bm{\theta}$, so that we can approximate the associated a posteriori probability $p(\bm{\theta}|\bm{w})$ with a simpler distribution $q(\bm{w})$, by maximizing a lower bound of the reverse KL-divergence $D\left(q(\bm{w})\|p(\bm{w}|\bm{\theta})\right)$. This indicates that the exact marginal probability of the noise $p(\bm{w})=\int p(\bm{w}, \bm{\theta}) \mathrm{d} \bm{\theta}$  as well as the likelihood of signal vectors remains computationally intractable.

\subsubsection{Data-Driven Methods}
In recent years, thanks to the tremendous success of deep learning, researchers have developed some data-driven methods to solve the problems encountered in various communication areas \cite{JiangDNC19,LichaoTVT,CuiLN20, XuYXLLC20}. For example, Y.-S. Jeon \textit{et al.} have proposed a supervised-learning based novel communication framework to construct a robust nonlinear MIMO system, which consisted of the concatenation of a wireless channel and a quantization function used at the ADCs for data detection \cite{JeonHL18}. For widely connected internet of things (IoT) devices, a novel deep learning-constructed joint transmission-recognition scheme was introduced in \cite{LeeLCC19} to tackle the crucial and challenging transmission and recognition problems.  It effectively improves the data transmission and recognition by jointly considering the transmission bandwidth, transmission reliability, complexity, and recognition accuracy. For the aspect of signal detection, the authors in \cite{samuel2019learning} proposed a projection gradient descent (PGD) based signal detection neural network (DetNet), by unfolding the iterative PGD process into a series of neural layers.  However, its performance is not guaranteed when the noise statistics is unknown, for which the gradient is computed based on the maximum likelihood criterion of Gaussian noise in DetNet. Moreover, when the noise is dynamically correlated in time or frequency domain, the authors proposed a deep learning based detection framework to improve the performance of MIMO detectors in \cite{XiaHXZFK20, HeWHDXF20}. In addition, some generative models based on deep learning have been proposed to learn the unknown distribution of random variables. In particular, the generative models are probabilistic, driven by unsupervised learning approaches. Currently, there are three major types of generative models \cite{Oussidi2018, Salih2014}, which are variational auto encoders (VAEs) \cite{kingma2014auto, rezende2015variational}, generative adversarial networks (GANs) \cite{Goodfellow2014Generative, Kurach2019} and normalizing flows (NFlows) \cite{dinh2014nice, Dinh2016Density, Diederik2018Glow}. Recently, the deep generative models are adopted in literature to solve the linear inverse problems \cite{Shirin2019Solving}. For example, the images can be restored with high quality from the noisy observations by approximating the natural distribution of images with a generative model  \cite{Jalali019}, which inspires us to try to solve the linear inverse problem with the aid of data-driven generative models rather than noise statistics.


\subsection{Contributions}
In this paper, we propose an effective MLE method to detect the signal, when the noise statistics is unknown. Specifically, we propose a novel signal detection framework, named \emph{maximum a normalizing flow estimate} (MANFE). This is a fully probabilistic model, which can efficiently perform MLE by approximating the unknown noise distribution through a normalizing flow. To reduce the computational complexity of MLE, we further devise a low-complexity version of the MANFE, namely G-GAMP-MANFE, by jointly integrating the G-GAMP algorithm and MANFE. The main contributions of this work can be summarized as follow:
\begin{itemize}
\item We propose a novel and effective MLE method, when  only noise samples are available rather than statistical knowledge. Experiments show that this method achieves much better performance than other relevant algorithms under impulsive noise environments. Also,  it can still reach the performance bound of MLE in Gaussian noise environments.
\item The proposed detection framework is very flexible, since it does not require any statistical knowledge on the noise. In addition, it is driven by an unsupervised learning approach, which does not need any labels for training.
\item The proposed detection framework is robust to  impulsive environments, since it performs better a more effective MLE with comparison compared to E-MLE, when the  noise  statistics is unknown.
\item We extend the MANFE by presenting a low-complexity version in order to reduce the computational complexity of MLE. The  complexity of this version is very low, so that it can be easily implemented in practical applications. Further experiments show that its performance can even outperform the E-MLE under highly impulsive noise environments.
\end{itemize}

\subsection{Organization}
In section \ref{section:flowreview}, we  first overview the prototype of normalizing flows and discuss the reasons why we choose normalizing flows to solve the problem under investigation.  The proposed detection framework and the implementation details are presented in Section \ref{sec:mafnet}. We present various simulation results and discussions in Section \ref{sec:experiments} to show the effectiveness of the proposed methods. Finally, we conclude the contribution of this paper in Section \ref{sec:conclusion}. 

\section{Unknown Distribution Approximation} \label{section:flowreview}
In this section, we firstly present the maximum likelihood approach for the distribution approximation, and then provide the concept of normalizing flows. Furthermore, we compare the normalizing flows with other generative models, and explain the reason why we choose the former method to approximate an unknown noise distribution.

\subsection{Maximum Likelihood Approximation}
Let $\bm{w}$ be a random vector with an unknown distribution $p_{\bm{w}}(\bm{w})$ and $\mathcal{D}_{\bm{w}} = \{ \bm{w}^{(1)}, \bm{w}^{(2)}, \cdots, \bm{w}^{(L)}\}$ is a collected data set consisting of $L$ i.i.d data samples. Using $\mathcal{D}_{\bm{w}}$, the  distribution $p_{\bm{w}}(\bm{w})$ can be approximated by maximizing the total likelihood of the data set on the selected model $q(\bm{w}; \bm{\theta})$ parameterized by $\bm{\theta}$, such as the mixture Gaussian models. In this case, the loss function is the sum of the negative log-likelihoods of the collected data set, which can be expressed as
\begin{align}\label{eq:objective}
\mathcal{L}(\bm{\theta}) = -\frac{1}{L}\sum_{l=1}^L \log q\left(\bm{w}^{(l)}; \bm{\theta}\right).
\end{align}
Clearly, (\ref{eq:objective}) measures how well the model $q(\bm{w}; \bm{\theta})$ fits the data set drawn from the distribution $p_{\bm{w}}(\bm{w})$. Since $q(\bm{w}; \bm{\theta})$ is a valid PDF, it is always nonnegative. In particular, it reaches its minimum if the selected model $q(\bm{w}; \bm{\theta})$ perfectly fits the data, i.e. $q(\bm{w}; \bm{\theta}) \equiv p_{\bm{w}}(\bm{w})$. Otherwise, it enlarges if $q(\bm{w}; \bm{\theta})$ deviates from $ p_{\bm{w}}(\bm{w})$. Hence, the training objective is to minimize the loss and find the optimal parameters as
\begin{align}
\bm{\theta}^* = \arg\min_{\bm{\theta}} \mathcal{L}(\bm{\theta}),
\end{align}
where $\bm{\theta}$ can be optimized by some methods, such as the stochastic gradient descent (SGD) with mini-batches of data \cite{Mu2014Efficient}. This is an unsupervised learning approach, since the objective does not require any labeled data. However, it is not flexible enough if the optimization is performed directly on the selected model $q(\bm{w}; \bm{\theta})$, since the knowledge of the true distribution is needed in order to choose an appropriate model for approximation. 

\subsection{Normalizing Flow}
As a generative model, \emph{normalizing flow} allows to perform efficient inference on the latent variables \cite{rezende2015variational}. More importantly, the computation of log-likelihood on the data set is accomplished by using the change of variable formula rather than computing on the model directly. For the observation $\bm{w} \in \mathcal{D}_{\bm{w}}$, it depends on a latent variable $\bm{z}$ whose density function $p(\bm{z}; \bm{\theta})$ is simple and computationally tractable (e.g. spherical multivariate Gaussian distribution), and we can describe the generative process as
\begin{align}
\bm{z} &\sim p(\bm{z}; \bm{\theta}), \\
\label{changeofvar}
\bm{w} &= g(\bm{z}),
\end{align}
where $g(\cdot)$ is an invertible function (aka bijection), so that we can infer the latent variables efficiently by applying the inversion $\bm{z}= f(\bm{w})  = g^{-1}(\bm{w}) $.  By using (\ref{changeofvar}), we can model the approximate distribution as
\begin{align}\label{eq:latentloglikelihood}
\log q(\bm{w}; \bm{\theta}) = \log p(\bm{z}; \bm{\theta}) + \log\bigg\vert \det\bigg(\frac{\mathrm{d} \bm{z}}{\mathrm{d} {\bm{w}}}\bigg) \bigg\vert,
\end{align}
where the so called \emph{log-determinant} term $\log\bigg\vert \det\bigg(\frac{\mathrm{d} \bm{z}}{\mathrm{d} {\bm{w}}}\bigg) \bigg\vert$ denotes the logarithm of the absolute value of the determinant on the Jacobian matrix $\left(\frac{\mathrm{d} \bm{z}}{\mathrm{d} \bm{w}}\right)$.  To improve the model flexibility, it is recognized that the invertible function $f(\cdot)$ is composed of $K$ invertible subfunctions
\begin{equation}
f(\cdot) = f_1(\cdot) \otimes f_2(\cdot) \otimes \cdots  f_k(\cdot) \cdots \otimes f_K(\cdot).
\end{equation}
From the above equation, we can infer the latent variables $\bm{z}$ by
\begin{align}\label{eq:latentmap}
\bm{w} \stackrel{f_1}{\longrightarrow} \bm{h_1} \stackrel{f_2}{\longrightarrow} \bm{h_2} \cdots \stackrel{f_k}{\longrightarrow} \bm{h}_k \cdots \stackrel{f_K}{\longrightarrow} \bm{z}.
\end{align}
By using the definitions of $\bm{h_0} \triangleq \bm{w}$ and $\bm{h_K} \triangleq \bm{z}$, we can rewrite the loss function in (\ref{eq:objective}) as
\begin{align}
\mathcal{L}(\bm{\theta}) &= -\frac{1}{L}\sum_{l=1}^L \log q(\bm{w}^{(l)}; \bm{\theta}) \\
&= -\frac{1}{L}\sum_{l=1}^L \left(\log p(\bm{z}^{(l)}; \bm{\theta}) + \sum_{k=1}^K \log\bigg\vert \det\bigg(\frac{\mathrm{d} \bm{h}^{(l)}_k}{\mathrm{d} \bm{h}^{(l)}_{k-1}}\bigg) \bigg\vert \right).
\end{align}
Using the above, we can treat each subfunction as a step of the flow, parameterized by trainable parameters. By putting all the $K$ flow steps  together, a normalizing flow is constructed to enable us to perform approximate inference and efficient computation of the log-probability.

In general, the normalizing flow is inspired by the change of variable technique. It assumes that the observed random variable comes from the invertible change of a latent variable which follows a specific distribution. Hence, the normalizing flow is actually an approximate representation of the invertible change. From this perspective, one simple example is the general Gaussian distribution. Let us treat the normalizing flow as a black box, and simply use $f(\cdot)$ to represent the revertible function parameterized by the normalizing flow. Since any Gaussian variable $X \sim \mathcal{N}(\mu, \sigma)$ can be derived via the change of the standard Gaussian variable $Y \sim \mathcal{N}(0, 1)$, the normalizing flow actually represents the approximation of the perfect inversion, saying that $Y = \frac{X-\mu}{\sigma} \approx f(X)$. In this case, one can easily compute the approximation of the corresponding latent variable as $Y \approx f(X)$. Therefore, when the observed variable follows different unknown distributions, the only difference is that the network parameters are fine tuned to different values for different distributions, which makes the principle of computing the associated latent variables become rather simple. To summarize the normalizing flow, its name has the following interpretations:
\begin{itemize}
\item ``Normalizing'' indicates that the density is normalized by the reversible function and the change of variables, and
\item ``Flow'' means that the reversible functions can be more complex by incorporating other invertible functions.
\end{itemize}

\subsection{Why to use the Normalizing Flow?}
Similar to normalizing flow, the other generative models like VAE and GAN map the observation into a latent space. However, the exact computation of log-likelihood is totally different in these models. Specifically, in the VAE model the latent variable is inferred by approximating the posterior distribution $p(\bm{z} | \bm{w})$. Hence, the exact log-likelihood can be computed through the marginal probability $p(\bm{w}) = \int p(\bm{w}, \bm{z}) \mathrm{d} \bm{z}$, with the help of numerical methods like Monte Carlo. Therefore, in this case the computational complexity of the log-likelihood is very high. On the other hand, the GAN model do not maximize the log-likelihood. Instead of the training via the maximum likelihood, it will train a generator and a discriminator. The generator maps the observation into a latent space and draws samples from the latent space, while the discriminator decides whether the samples drawn from the generator fit the collected data set. Hence, both generator and discriminator play a min-max game. In this case, drawing samples from GANs is easy, while the exact computation of the log-likelihood is computationally intractable.

For the reversible generative models like normalizing flows, the inference of latent variables can be exactly computed. The benefit of this approach is that we are able to compute the corresponding log-likelihood efficiently. In consequence, the normalizing flow is a good choice among these three generative models to perform exact Bayesian inference, especially when the noise statistics is unknown.

\section{Proposed Signal Detection Framework}\label{sec:mafnet}
In this section, we propose an unsupervised learning driven and normalizing flow based signal detection framework, which enables the effective and fast evaluation of MLE without knowledge of the noise statistics.

\begin{figure}[t!]
    \centering
    \includegraphics[height=4in]{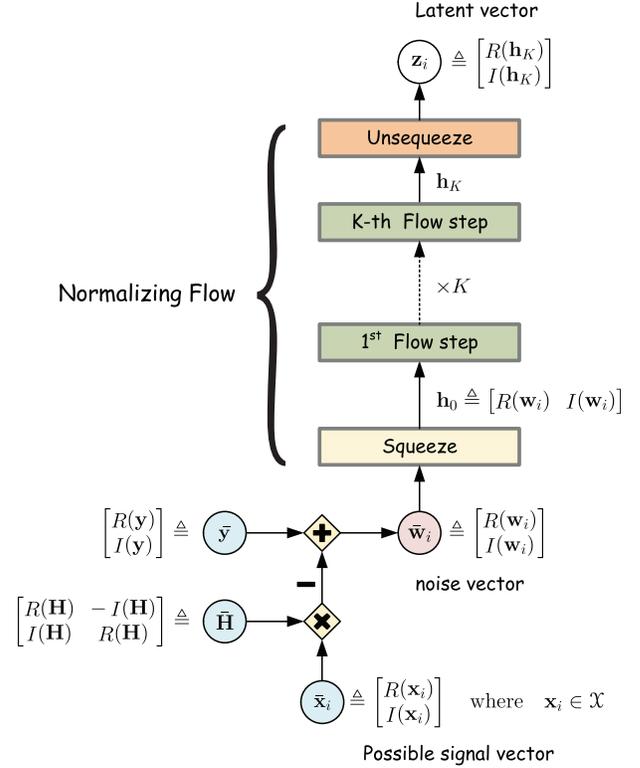}
    \caption{Architecture of the detection framework with normalizing flow.}
    \label{fig:flownet}
\end{figure}

As shown in Fig. \ref{fig:flownet}, the proposed detection framework is built upon a normalizing flow, which  includes three kinds of components:  squeeze layer,  $K$ flow steps, and an unsqueeze layer. To perform the MLE, given $\bm{y}$ and $\bm{H}$, we firstly compute the associated noise vector $\bm{w}_i = \bm{y} - \bm{H}\bm{x}_i$ for each possible signal vector $\bm{x}_i \in \mathcal{X}$. Then, by using the normalizing flow, we can map $\bm{w}_i$ into the latent space, and infer the associated latent variable $\bm{z}_i$  as well as the log-determinant. Therefore we can compute the corresponding log-likelihood $p(\bm{y} | \bm{x}_i)$ from (\ref{eq:latentloglikelihood}) and determine the final estimation of the maximum log-likelihood.

Since most neural networks focus on real-valued input, we can use a well-known real-valued representation to express the complex-valued input equivalently as
\begin{align}
\label{eq:realpresentation}
\underbrace{\begin{bmatrix}
\mathop{R}(\bm{y}) \\
\mathop{I}(\bm{y}) \\
\end{bmatrix}}_{\bar{\bm{y}}}
=
\underbrace{\begin{bmatrix}
\mathop{R}(\bm{H}) & -\mathop{I}(\bm{H}) \\
\mathop{I}(\bm{H}) & \mathop{R}(\bm{H}) \\
\end{bmatrix}}_{\bar{\bm{H}}}
\underbrace{\begin{bmatrix}
\mathop{R}(\bm{x}) \\
\mathop{I}(\bm{x}) \\
\end{bmatrix}}_{\bar{\bm{x}}}
+
\underbrace{\begin{bmatrix}
\mathop{R}(\bm{w}) \\
\mathop{I}(\bm{w})
\end{bmatrix}}_{\bar{\bm{w}}},
\end{align}
where $\mathop{R}(\cdot)$ and $\mathop{I}(\cdot)$ denote the real and imaginary part of the input, respectively. Then, we can rewrite (\ref{eq:realpresentation}) into a form like (\ref{eq:linearmodel}) as
\begin{align}
\bar{\bm{y}} = \bar{\bm{H}}\bar{\bm{x}}+\bar{\bm{w}}.
\end{align}
Based on the above real-valued representation, the squeeze layer will process the complex-valued input by separating the real and imaginary part from the input, while the unsqueeze layer performs inverselly. In general, an $M$-dimensional complex-valued input $\bm{h}=\begin{bmatrix} h_1, h_2, \cdots, h_M \end{bmatrix}$ for a flow step can be represented by a 2-D tensor with a shape of $M \times 2$ and the following structure
\begin{align}
    \begin{bmatrix}
         \mathop{R}(h_1) & \mathop{I}(h_1)  \\
         \mathop{R}(h_2) & \mathop{I}(h_2)  \\
         \vdots & \vdots \\
         \mathop{R}(h_M) & \mathop{I}(h_M)  \\
    \end{bmatrix},
\end{align}
where the first column/channel includes the real  and the second column/channel the imaginary part, respectively.

\begin{figure}[t!]
    \centering
    \includegraphics[height=4in]{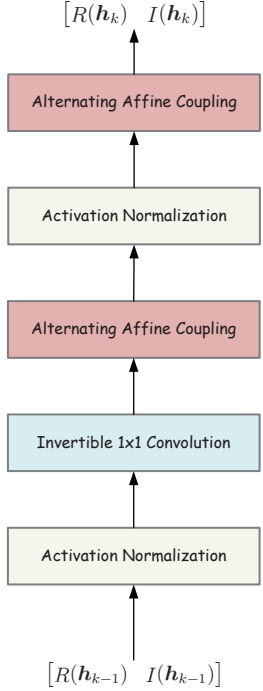}
    \caption{Structure of one step of the normalizing flow. }
    \label{fig:flowstep}
\end{figure}

From Section \ref{section:flowreview} we can conclude that a critical point when implementing a normalizing flow is to carefully design the architecture,  in order to ensure that the subfunctions, represented by neural blocks, are invertible and flexible, and the corresponding log-determinants are computationally tractable. Accordingly, as shown in Fig. \ref{fig:flowstep}, each flow step consists of three hidden layers: an activation normalization, an invertible $1 \times 1$ convolution layer, and an alternating affine coupling layer. In particular, all these hidden layers except squeeze and unsequeeze, would not change the shape of input tensor. This means that its output tensor will have the same structure as the input one. In the rest of this section, we introduce these hidden layers one by one and then we present how the detection framework works.

\subsection{Activation Normalization}
To accelerate the training of the deep neural network, we adopt a batch normalization technique \cite{BatchNormalization} and employ an activation normalization layer \cite{Diederik2018Glow} to perform inversely affine transformation of activations. The activation normalization at the $k$-th layer utilizes trainable scale and bias parameters for each channel, and the corresponding initial value depends on the initial mini-batch of input data $\bm{h}_{k-1}$ from the prior layer
\begin{align}\label{eq:actnorm}
\bm{s}_k &= \frac{1}{\sqrt{\mathbb{V}\left[\bm{h}_{k-1}\right]}}, \\
\bm{b}_k &= -\mathbb{E}\left[\bm{h}_{k-1}\right],
\end{align}
where $\mathbb{E}[\cdot]$ and $\mathbb{V}[\cdot]$ denote the expectation and variance, respectively.
Once the initialization is performed,  the scale and bias are treated as trainable parameters and then the activation normalization layer output is
\begin{align}\label{eq:actnorm_f}
\bm{h}_k = \bm{h}_{k-1} \odot \bm{s}_k +  \bm{b}_k,
\end{align}
with $\odot$ being the element-wise multiplication on the channel axis. As explained in \cite{Dinh2016Density} and \cite{Diederik2018Glow}, the Jacobian of the transformation of coupling layers depends on the associated diagonal matrix. Due to that the determinant of a triangular matrix is equal to the product of the diagonal elements, the corresponding log-determinant can be easily computed. Specifically, the log-determinant of (\ref{eq:actnorm_f}) is given by
\begin{align}
\log \left\vert \det\left(\frac{\mathrm{d} \bm{h}_{k}}{\mathrm{d} \bm{h}_{k-1}}\right) \right \vert = M \text{sum} \left( \log \lvert \sqrt{\bm{s}_k}\rvert \right),
\end{align}
where the operator $\text{sum}(\cdot)$ means sum over all the elements of a tensor.

\subsection{Invertible $1 \times 1$ Convolution}
In order to improve the model flexibility, we employ an invertible $1 \times 1$ convolutional layer. This can  incorporate the permutation into the deep neural network, while it does not change the channel size and it can be treated as a generalized permutation operation \cite{Diederik2018Glow}. For an invertible $1 \times 1$ convolution with a $2 \times 2$ learnable weight matrix $\bm{W}_k$, the log-determinant is computed directly as  \cite{Diederik2018Glow}
\begin{align}\label{eq:1x1invert}
\log \left\vert \det\left(\frac{\mathrm{d} \bm{h}_{k}}{\mathrm{d} \bm{h}_{k-1}}\right) \right \vert = M \log \lvert \det(\bm{W}_k)\rvert.
\end{align}
In order to construct an identity transformation at the beginning, we set the weight $\bm{W}_k$ to be a random rotation matrix with a zero log-determinant at initialization.

\subsection{Alternating Affine Coupling Layer}
Affine coupling layer is a powerful and learnable reversible function, since  its log-determinant is computationally tractable\cite{dinh2014nice,Dinh2016Density}. For an $M$-dimensional input $\bm{h}_{k-1}$, the affine coupling layer will change some part of the input based on another part of the input. To do this, we first separate the input into two parts, which is given by
\begin{subequations}
\begin{align}
\bm{q}_k &= \bm{h}_{k-1}(1 \colon m), \\
\bm{u}_k &= \bm{h}_{k-1}(m+1 \colon M),
\end{align}\label{eq:affine_coupling}
where $\bm{h}_{k-1}(1 \colon m)$ represents the first part of the input,  and $\bm{h}_{k-1}(m+1 \colon M)$ denotes the rest. After that, we couple the two parts together in some order
 \begin{align}
\bm{s}_k &= \mathcal{G}(\bm{q_k}) \quad \mbox{or} \quad \mathcal{G}(\bm{u_k}), \\
\bm{b}_k &= \mathcal{H}(\bm{q_k}) \quad \mbox{or} \quad \mathcal{H}(\bm{u_k}), \\
\bm{h}_k &= \bm{h}_{k-1}, \\
\bm{h}_k({m+1\colon M}) &= \bm{u}_k \odot \bm{s}_k + \bm{b}_k,
\end{align}\label{eq:affine_coupling}
\end{subequations}
$\mathcal{G}(\cdot) $ and $\mathcal{H}(\cdot) $  represent two learnable neural networks, which dynamically and nonlinearly compute the corresponding scale and bias. Similarly, the determinant of a triangular matrix is again equal to the product of the diagonal elements \cite{Dinh2016Density}, which indicates that we can readily compute the corresponding log-determinant as
\begin{align}
\log \left\vert \det\left(\frac{\mathrm{d} \bm{h}_{k}}{\mathrm{d} \bm{h}_{k-1}}\right) \right \vert = \text{sum} \left( \log \lvert \bm{s}_k \rvert \right).
\end{align}

Basically, the architecture of our flow steps is developed based on the implementations suggested in \cite{dinh2014nice, Dinh2016Density, Diederik2018Glow}. However, there are two key differences between our architecture and them. One key difference is that our implementation can handle the complex-valued input by using the squeeze layer and the unsqueeze layer, in order to separate and recover the real parts and imaginary parts. Another key difference is that we just need to ensure the existence of inversions rather than investigating their exact analytical forms, since there is no need to draw samples from the latent space in signal detection. This difference leads to more generalities and flexibilities enabled in our implementation. Specifically, as shown in Fig. 2, we alternatively combine two affine coupling layers together in a flow step, and they can change any part of the input without considering whether it has been changed or not at the prior coupling layer. As a contrast, the existing implementations suggested in \cite{dinh2014nice, Dinh2016Density, Diederik2018Glow} must change the part that has not been changed yet by the prior layer. Therefore, we incorporate a powerful alternating pattern into the network to help enhance the generality and flexibility, and eventually improve the network's ability to approach unknown distributions.

\subsection{Signal Detection}\label{sec:manfe}
\begin{algorithm}[t]
	\caption{MANFE}
    \algorithmfootnote{$f(\cdot)$ denotes an invertible function  represented by a normalizing flow and $\mathcal{X}$ denotes  a set which contains all possible signal candidates.}
	\label{algo:manfe}
	\LinesNumbered
	\KwIn{Received signal $\bm{y}$, channel matrix $\bm{H}$,}
	\KwOut{Recovered signal $\bm{x}^*$}
	\ForEach{$\bm{x}_i \in \mathcal{X}$}{
        $\bm{w}_i \gets \bm{y}-\bm{H}\bm{x}_i$ \\
        $\bm{z}_i \gets f(\bm{w}_i) $ \\
        $\mathcal{L}_{\bm{x}_i} \gets \log p(\bm{z}_i; \bm{\theta}) + \log\left\vert \det\left(\frac{\mathrm{d} \bm{z}_i}{\mathrm{d} \bm{w}_i}\right) \right\vert$
    }
    $\bm{x}^* \gets \arg\underset{\bm{x}_i}{\max}(\mathcal{L}_{\bm{x}_i})$ \\
    \Return $\bm{x}^*$
\end{algorithm}

As discussed above, by jointly using the change of latent variables and the nonlinearity of neural networks, the approximate distribution parameterized by a normalizing flow can be highly flexible to approach the unknown true distribution. In this case, we are able to perform MLE in (\ref{eq:map}) by evaluating the log-likelihood through the normalizing flow. Accordingly, we devise a signal detection algorithm for the proposed detection framework. In contrast to E-MLE, the proposed detection framework estimates the signal by finding the maximum likelihood computed from a normalizing flow, so that we call it \emph{maximum a normalizing flow estimate} (MANFE).

Specifically, in the MANFE algorithm, we first evaluate the corresponding noise vector $\bm{w}_i$ given received signal $\bm{y}$ and channel maxtrix $\bm{H}$ for each possible signal vector $\bm{x}_i \in \mathcal{X}$. Then, the algorithm maps the noise vector $\bm{w}_i$ into the latent space to infer the corresponding latent variable $\bm{z}_i$. After that, we compute the corresponding log-likelihood by evaluating
\begin{align}
\mathcal{L}_{\bm{x}_i} &= \log p(\bm{y}|\bm{x}_i)  \\
&\approx \log q(\bm{w}_i; \bm{\theta}) \\
&= \log p(\bm{z}_i; \bm{\theta}) + \log\left\vert \det\left(\frac{\mathrm{d} \bm{z}_i}{\mathrm{d} \bm{w}_i}\right) \right\vert.
\end{align}
Finally, by finding the most possible signal vector which has the maximum log-likelihood, we get the MLE of desired signal as
\begin{align}
\bm{x}^* = \arg\underset{\bm{x}_i \in \mathcal{X}}{\max}(\mathcal{L}_{\bm{x}_i}).
\end{align}
The whole procedures of the MANFE algorithm is summarized in Algorithm \ref{algo:manfe}. Intuitively, the major difference between MANFE and E-MLE is that MANFE is a generalized ML estimator which approximates the unknown noise distributions and thereby compute the log-likelihoods accordingly in different noise environments. On the contrary, E-MLE is only designed for Gaussian noises so that it can be treated as a perfectly trained MANFE under specific noise environment, which results that it loses flexibilities with comparison to MANFE.

\subsection{Low-Complexity Version of MANFE}\label{sec:lowmanfe}
As the MLE is an NP-hard problem, we have to exhaust all possible candidates to check out which one has the maximum probability. In particular, the computational complexity of MLE is $\mathcal{O}(P^N)$, which indicates that the complexity increases exponentially with the constellation size $P$ and the antenna number $N$. Hence, it is difficult to implement a perfect MLE in practice.

To solve this problem, some empirical approaches have been proposed to reduce the complexity of MLE, by utilizing an initial guess to reduce the searching space \cite{Pammer2003A}. The initial guess can be estimated by some kind of low-complexity detectors, such as ZF detector, MMSE detector, and GAMP detector. Though the selection range of low-complexity detectors is broadly wide, we should choose a detector which has a lower complexity and fine bit error rate (BER) performance. From this viewpoint and in order to reduce the complexity of MLE, we propose to jointly use the low-complexity G-GAMP detection algorithm and the MAFNE, which is named as G-GAMP-MANFE.

Specifically, in the G-GAMP-MANFE algorithm, we first get an initial estimate came from G-GAMP algorithm, where the details about the GAMP algorithm can be found in the existing works such as \cite{donoho2009message, Rangan2010Generalized, rangan2016fixed}. Since the initial guess is approximate, we can assume that there exist at most $E$ ($0 \leq E \leq N$) error symbols at the initial guess. Accordingly, we will only require to compare $\sum_{i=0}^E C_{N}^i (P-1)^{i}$ signal candidates instead of $P^N$ ones. The number of error symbols can be sufficiently small so that the total complexity can be reduced significantly, especially when the channel is in good condition. For example, we only need to compare $1+N(P-1)$ candidates when $E = 1$. Hence, the searching space is reduced as well as the total complexity of MANFE.  The whole procedures of the G-GAMP-MANFE algorithm is summarized in Algorithm \ref{algo:gampmanfe}, as shown at the top of the next page.

Basically, the low-complexity version of MANFE is a generalized framework that helps improve the detection performance of other low-complexity detectors under unknown noise environments. In this paper, we use G-GAMP detector for the initial estimation as its performance and complexity are both acceptable under most common scenarios \cite{LiuLHYG19}. Obviously, the BER performance of the low-complexity MANFE depends on two factors. One is that the choice of initial detector significantly affects the BER performance of the low-complexity MANFE, while the other is the choice of $E$. If the problem scale is not too large and we can tolerate a high complexity, we can increase $E$ to improve the BER performance. On the other hand, if the systems are sensitive to the computational complexity, it would be better to maintain $E$ at a lower level. In other worlds, the choice of $E$ is quite flexible and users would set the value of $E$ based on their specific needs. In particular, as an iterative detection algorithm, G-GAMP-MANFE's convergence mainly depends on the G-GAMP, whose convergence can be guaranteed when the noise is Gaussian and the channel matrix is a large i.i.d. sub-Gaussian matrix, where the details can be found in the literature such as \cite{Adel2013State, rangan2016fixed, liu2019capacity}.

\begin{algorithm}[t]
	\caption{Combine G-GAMP With MANFE (G-GAMP-MANFE)}
    \algorithmfootnote{$f(\cdot)$ denotes an invertible function  represented by a normalizing flow and $\mathcal{X}$ denotes a set which contains all possible signal candidates.}
	\label{algo:gampmanfe}
	\LinesNumbered
	\KwIn{Received signal $\bm{y}$, channel matrix $\bm{H}$}
	\KwOut{Recovered signal $\bm{x}^*$}
    Get an initial estimate $\bm{x}_0$ from G-GAMP algorithm \\
    Get a subset $\mathcal{X}_E$ from $\mathcal{X}$ where there exist at most  $E$ different symbols between a possible signal candidate and the initial estimate $\bm{x}_0$ \\
	\ForEach{$\bm{x}_i \in \mathcal{X}_E$}{
        $\bm{w}_i \gets \bm{y}-\bm{H}\bm{x}_i$ \\
        $\bm{z}_i \gets f(\bm{w}_i) $ \\
        $\mathcal{L}_{\bm{x}_i} \gets \log p(\bm{z}_i; \bm{\theta}) + \log\left\vert \det\left(\frac{\mathrm{d} \bm{z}_i}{\mathrm{d} \bm{w}_i}\right) \right\vert$
    }
    $\bm{x}^* \gets \arg\underset{\bm{x}_i}{\max}(\mathcal{L}_{\bm{x}_i})$ \\
    \Return $\bm{x}^*$
\end{algorithm}

\subsection{Complexity Analysis}\label{sec:complexity}
In this part, we provide some analysis on the computational complexity for the MANFE and G-GAMP-MANFE algorithms. There are three kinds of subfunction in the MANFE detection framework and all these subfunctions operate in an element-wise manner. Accordingly, the computational complexity of a flow step depends on the element-wise summation of log-determinant, which is about $\mathcal{O}(M)$ for a single flow step. Therefore, the computational cost to compute the log-likelihood of a possible signal vector $\bm{x}_i$  depends on the matrix multiplication $\bm{H}\bm{x}_i$ and the cost of $K$ flow steps, which is about $\mathcal{O}(KM+MN)$. Hence, the total computational complexity for MANFE can be expressed as $\mathcal{O}((KM+MN)P^N)$, for which we have to exhaust all $P^N$ possible signal candidates.

As to the G-GAMP-MANFE algorithm, since the G-GAMP has a computational complexity of $\mathcal{O}(T(M+NP))$ for $T$ iterations and we have to compare $\sum_{i=0}^E C_{N}^i (P-1)^{i}$ candidates, the computational complexity for the G-GAMP-MANFE is $\mathcal{O}\big(T(M+NP) + (KM+MN) \sum_{i=0}^E C_{N}^i (P-1)^{i}\big)$. More specifically, when $E=1$, the complexity of the G-GAMP-MANFE is only of $\mathcal{O}\big( T(M+NP) + (KM+MN)(1+N(P-1)) \big)$, which indicates that it can be easily implemented in practical MIMO systems.

\section{Simulations and Discussions}\label{sec:experiments}
In this section, we perform simulations to verify the effectiveness of the proposed detection framework. In particular, we first introduce the environment setup of these simulations as well as the implementation details of the deep neural network, and then, we present some simulation results and give the related discussions.

\subsection{Environment Setup}\label{sec:environmentsetup}
The simulations are performed in a MIMO communication system in the presence of several typical additive non-Gaussian noises, such as Gaussian mixture noise, Nakagami-$m$ noise, and impulsive noise. The numbers of antennas are $N$ and $M$ at the transmitter and the receiver, respectively. The modulation scheme is quadrature phase shift keying (QPSK) with $P=4$, and the channel experiences Rayleigh flat fading. The receiver has perfect knowledge on the channel state information (CSI). To model the impulsive noise, we employ a typical impulsive model, named symmetric $\alpha$-stable (S$\alpha$S) noise\cite{Samorodnitsky1996Stable, Fan2012On, Chen2015Suboptimum, Chen2015Novel}. In particular, an S$\alpha$S random variable $w$ has the following characteristic function
\begin{align}
\psi_w(\theta) = \mathbb{E}[e^{jw \theta}] = e^{-\sigma^{\alpha} \lvert \theta \rvert^\alpha},
\end{align}
where $\mathbb{E}[\cdot]$ represents the statistical expectation, $\sigma > 0$ is the scale exponent, and $\alpha \in (0, 2]$ is the characteristic exponent. When $\alpha$ decreases, the noise becomes heavy-tailed and impulsive. For practical scenarios, $\alpha$ usually falls into $[1, 2]$. Especially, the S$\alpha$S distribution turns into a Cauchy distribution when $\alpha=1$, while it is a Gaussian distribution when $\alpha=2$. The density function  $f_w(w)$ can be expressed by \cite{Samorodnitsky1996Stable}
\begin{align}
\label{eq:salphasnoise}
f_w(w) = \frac{1}{2\pi} \int_{-\infty}^{+\infty} e ^ {-\lvert \theta \rvert \sigma^\alpha - j\theta w } \mathrm{d} \theta.
\end{align}

Unfortunately, we can only compute the approximate density through numerical methods since $f_w(w) $ does not have a closed-form expression when $\alpha \in (1, 2)$. In other words, the exact MLE under S$\alpha$S noise is computationally intractable and the performance of E-MLE will severely deviate from the situation of Gaussian noise. In particular, we mix two Gaussian distributed noises subject to $\mathcal{CN}(-\mathbf{I}, 2\mathbf{I})$ and $\mathcal{CN}(\mathbf{I}, \mathbf{I})$ equably as the instance of the Gaussian mixture noise. Notice that any statistical knowledge on these noises such as the value of $\alpha$ which indicates the impulse level is not utilized during the training and testing processes, which can simulate the situation that the noise statistics is unknown.

\subsection{Training Details}
In the proposed framework, the hyper-parameters that need to be chosen carefully are concluded as follows:
 \begin{itemize}
 \item The total number of flow steps $K$,
 \item The specified partition parameter $m$ in alternative affine coupling layers,
  \item The specified structure of the two neural networks $\mathcal{G}(\cdot)$ and $\mathcal{H}(\cdot)$ in alternative affine coupling layers.
 \end{itemize}
Obviously, $K$ significantly affects the complexity and the effectiveness of our methods, and we find that $K=4$ is a good choice based on the experiences.  As $\bm{q}_k$ is often the half part of the input $\bm{h}_{k-1}$, we can set the partition parameter $m$ to $\frac{M}{2}$. Additionally, the scale and bias functions $\mathcal{G}(\cdot)$ and $\mathcal{H}(\cdot)$ are both implemented by a neural network with three fully-connected layers, where the activation functions are rectified linear unit (ReLU) functions and the hidden layer sizes are all constantly set to $8$  for both $4 \times 4$ and $8 \times 8$ MIMO systems.

In practice, we consider that the latent variables follow a multivariate Gaussian distribution with trainable mean $\bm{\mu}$ and variance $\bm{\Sigma}$. Therefore, the trainable parameters of the proposed framework can be summarized below:
\begin{itemize}
\item The scale vectors $\bm{s}_k$ and bias vectors $\bm{b}_k$ for each activation normalization layers introduced in (\ref{eq:actnorm}),
\item The learnable weight matrix $\bm{W}_k$ for each $1 \times 1$ convolution layer introduced in (\ref{eq:1x1invert}),
\item The network parameters in $\mathcal{G}(\cdot)$ and $\mathcal{H}(\cdot)$ for each affine coupling layer introduced in (\ref{eq:affine_coupling}),
\item The mean $\bm{\mu}$ and variance $\bm{\Sigma}$ of the latent variable's multivariate Gaussian distribution.
\end{itemize}
Hence, we can find that there are only a handful of trainable parameters inside a single flow step and the total number of flow steps is small too. This indicates that the training overhead of our framework is affordable. Specifically, we use the TensorFlow framework \cite{abadi2016tensorflow} to train and test the model, and there are $10$ millions training noise samples and $2$ millions test noise samples generated from the aforementioned noise models in the training phase. Since our model is driven by unsupervised learning, the data sets only consist of noise samples. To refine the trainable parameters, (\ref{eq:objective}) is adopted as the loss function, and the Adam optimizer \cite{kingma2015adam} with learning rate set to $0.001$ is adopted to update the trainable parameters by using the gradient descent method. Furthermore, engineering details and reproducible implementations can be found in our open-source codes located at \url{https://github.com/skypitcher/manfe} for interested readers.

\begin{table*}[t!]
\centering
\caption{Computational Complexity of Competing Algorithms\label{tab:complexity}}
\begin{tabular}{ll}
\toprule
{\bf \small Abbreviations} &\qquad {\bf\small Complexity}\\
\midrule
G-GAMP($T$)  & $\mathcal{O}\left(T(N+M)\right)$ \\
DetNet($T$)       & $\mathcal{O}\left(T(N^2+(3N+2L_z)L_h)\right)$ \footnotemark \\
E-MLE     & $\mathcal{O}\left(MNP^N\right)$  \\
MANFE  & $\mathcal{O}\left((KM+MN)P^N\right)$  \\
G-GAMP($T$)-MANFE($E$) & $\mathcal{O}\left(T(M+NP) + (KM+MN) \sum_{i=0}^E C_{N}^i (P-1)^{i}\right)$  \\
\bottomrule
\end{tabular}
\end{table*}
\footnotetext{$L_z$ and $L_h$ are the size of a latent parameter and the size of hidden layers, which are both suggested to be $2N$ in the paper of DetNet \cite{samuel2019learning}, respectively.}

\begin{table*}[t!]
\centering
\caption{Computational Complexity of Competing Algorithms in $4\times 4$ QPSK Modulated MIMO Systems\label{tab:time}}
\begin{tabular}{lccccc}
\toprule
{\bf \small Algorithm} &\qquad {\bf\small Complexity}\\
\midrule \vspace{1mm}
G-GAMP($30$)            & $\mathcal{O}\left(480\right)$ \\
DetNet($30$)            & $\mathcal{O}(28800)$ \\
E-MLE                   & $\mathcal{O}(16384)$  \\
MANFE                   & $\mathcal{O}(24576)$  \\
G-GAMP($30$)-MANFE($2$) & $\mathcal{O}(7152)$  \\
\bottomrule
\end{tabular}
\end{table*}

\subsection{Comparison with Relevant Algorithms}
In order to verify the effectiveness of the proposed detection framework, we compare the proposed methods with various competitive algorithms. For convenience,  we use the following abbreviations,
\begin{itemize}
\item \textbf{G-GAMP($T$)}: Gaussian GAMP algorithm with $T$ iterations. See \cite{donoho2009message, Rangan2010Generalized, rangan2016fixed} for more details.
\item \textbf{DetNet($T$)}: Deep learning driven and projected gradient descent (PGD) based detection network introduced in \cite{samuel2019learning} with $T$ layers (iterations).
\item \textbf{E-MLE}: Euclidean distance based maximum likelihood estimate introduced in (\ref{eq:mld}).
\item \textbf{MANFE}: Maximum a normalizing flow estimate introduced in Section \ref{sec:manfe}.
\item \textbf{G-GAMP($T$)-MANFE($E$)}: A low-complexity version of MANFE combined by the G-GAMP algorithm with $T$ iterations and the assumption that there exist at most $E$ error symbols at the initial guess came from the G-GAMP algorithm. See Section \ref{sec:manfe} and \ref{sec:lowmanfe} for more information.
\item \textbf{G-GAMP($T$)-E-MLE($E$)}: As similar to the G-GAMP-MANFE, the G-GAMP-E-MLE is a low-complexity version of E-MLE combined by the G-GAMP algorithm with $T$ iterations and the assumption that there are at most $E$ error symbols at the initial guess came from the G-GAMP algorithm.
\end{itemize}

For the purpose of complexity comparison, we provide two tables to present the complexity comparison among the competing algorithms. Specifically, Table \ref{tab:complexity} lists the theoretical computational complexity of the competing algorithms, while Table \ref{tab:time} presents the corresponding complexities in $4 \times 4$ QPSK modulated MIMO systems. From these two tables, we can find that although the computational complexity of the proposed MANFE is slightly higher than that of the conventional E-MLE, the complexity of the proposed G-GAMP-MANFE is affordable with a fine detection performance.

\subsection{Simulation Results and Discussions}
\begin{figure}[t!]
    \centering
    \includegraphics[width=3in]{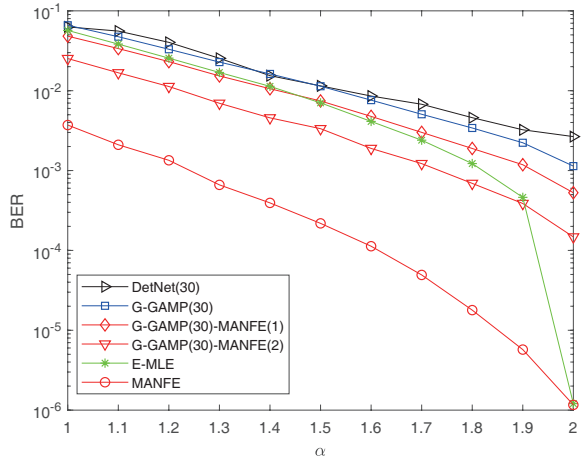}
    \caption{BER performance comparison versus $\alpha$ with SNR=25 dB for $4 \times 4$ MIMO systems.}\label{fig:bervsalpha}
\end{figure}

Fig. \ref{fig:bervsalpha} demonstrates the detection BER performances of the aforementioned detection methods, where the QPSK modulated $4 \times 4$ MIMO system is used with SNR$=25$ dB and $\alpha$ varies from $1$ to $2$. In particular, $\alpha=1$ and $\alpha=2$ correspond to the typical Cauchy and Gaussian distributed noise, respectively. We can find from Fig. \ref{fig:bervsalpha} that the proposed MANFE outerperforms the other several detectors in the impulsive noise environments, in the terms of BER performance. Specifically, when $\alpha=1.9$ where the noise is slightly impulsive and deviates a little from the Gaussian distribution, the MANFE can significantly reduce the detection error of E-MLE to about $1.1$\%. This indicates that the MANFE can compute an effective log-likelihood with contrast to E-MLE, especially when the noise is non-Gaussian with unknown statistics. More importantly, when $\alpha$ changes from $2.0$ to $1.9$ associated with a little impulsiveness, the E-MLE meets a severe performance degradation, whereas the MANFE has a relatively slight performance dagradation. This indicates that the MANFE is robust to the impulsive noise compared with E-MLE. In further, when the impulsive noise approaches to the Gaussian distribution with $\alpha=2$, the MANFE achieves the performance bound of MLE, indicating that the MANFE is very flexible to approach unknown noise distribution even when the noise statistics is unknown.

In addition, we can observe from Fig. \ref{fig:bervsalpha} that as a low-complexity version of MANFE, the G-GAMP-MANFE, still outperforms the other low-complexity detectors in impulsive noise environments. In particular, the G-GAMP(30)-MANFE(1) can even achieve the same performance as the E-MLE when $\alpha \leq 1.7$. In these cases, the G-GAMP(30)-MANFE(1) has a much lower computational complexity with comparison to the E-MLE, which is only about $5.08\%$\footnote{When $E=1$, the G-GAMP(30)-MANFE(1) can reduce the computational complexity of the E-MLE to about $\frac{1+N(P-1)}{P^N}$, which is equal to $\frac{13}{256} \approx 5.08\%$.} of the E-MLE. When $\alpha \le 1.9$, we can find that if we can tolerate a moderate computational complexity by increasing $E$ from $1$ to $2$,  the G-GAMP(30)-MANFE(2) will have a better BER performance than the E-MLE, which reduces the detection error of the E-MLE to about $47.5\%$ and meanwhile decreases the computational complexity to about $26.17\%$\footnote{Similarly, when $E=2$, the G-GAMP(30)-MANFE(2) can reduce the computational complexity of the E-MLE to about $\frac{\sum_{i=0}^2 C_{N}^i (P-1)^{i}}{P^N}$, which is equal to $\frac{67}{256} \approx 26.17\%$.} with respect to the E-MLE. By comparing with the other low-complexity detectors, the G-GAMP(30)-MANFE(1) can reduce the detection error of the DetNet(30) and the G-GAMP(30) to about $39.61\%$ and  $57.55\%$, respectively, when $\alpha=1.9$. These results further verify the effectiveness of the proposed detection framework.

\begin{figure}[t!]
    \centering
    \includegraphics[width=3in]{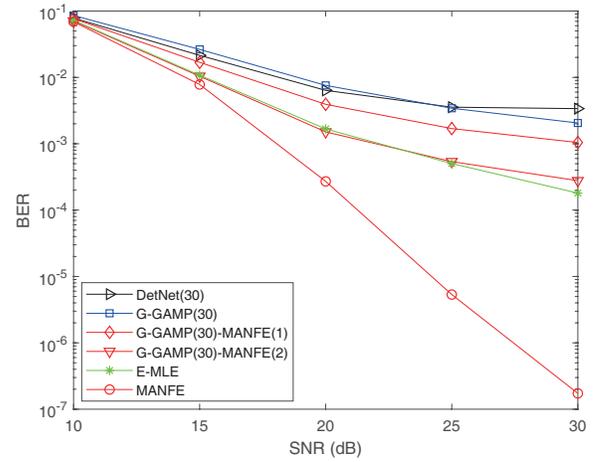}
    \caption{BER performance comparison versus SNR with $\alpha=1.9$ for $4 \times 4$ MIMO systems. }\label{fig:bervsnr1}
\end{figure}

\begin{figure}[t!]
    \centering
    \includegraphics[width=3in]{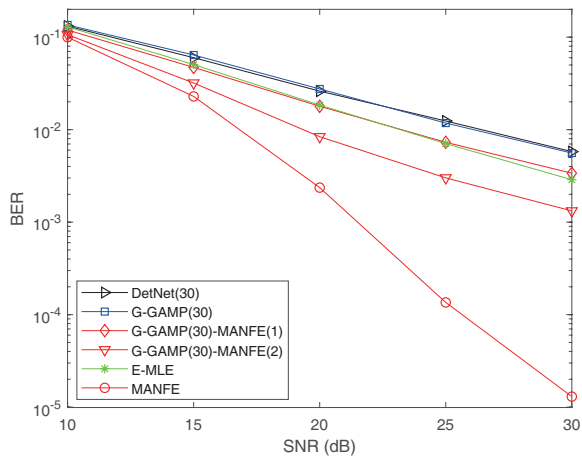}
    \caption{BER performance comparison versus SNR with $\alpha=1.5$ for $4 \times 4$ MIMO systems.}\label{fig:bervsnr2}
\end{figure}

\begin{figure}[t!]
    \centering
    \includegraphics[width=3in]{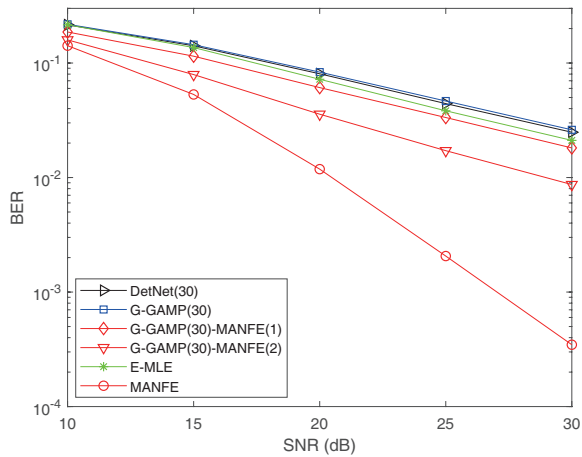}
    \caption{BER performance comparison versus SNR with $\alpha=1.1$ for $4 \times 4$ MIMO systems.}\label{fig:bervsnr3}
\end{figure}

To verify the effectiveness of the proposed detection framework under different channel conditions and different levels of impulsive noise,  Figs. \ref{fig:bervsnr1}-\ref{fig:bervsnr3} illustrate the BER performance comparison among several detectors for $4 \times 4$ MIMO systems in impulsive noise environments, where SNR varies from $10$ dB to $30$ dB. Specifically, Fig. \ref{fig:bervsnr1}, Fig. \ref{fig:bervsnr2} and Fig. \ref{fig:bervsnr3} are associated with $\alpha=1.9$, $\alpha=1.5$ and $\alpha=1.1$, respectively. From Figs. \ref{fig:bervsnr1}-\ref{fig:bervsnr3}, we can find that the BER performance gap between the MANFE and the E-MLE enlarges when the corresponding SNR increases. For example, when $\alpha=1.9$ and the values of SNR are set to $20$ dB, $25$ dB and $30$ dB, the MANFE can reduce the detection error of E-MLE to about $16.18$\%, $1.06$\% and $0.1$\%, respectively. In particular, for $\alpha=1.9$ when the noise is slightly impulsive, the SNR gain of the MANFE over the E-MLE is $10$ dB at the BER level of $10^{-4}$. Similarly, when the noises have moderate and strong level of impulsiveness where the corresponding values of $\alpha$ are equal to $1.5$ and $1.1$, the SNR gains of the MANFE over the E-MLE are about $10.5$ dB and $12$ dB at the BER levels of $10^{-3}$ and $10^{-2}$, respectively. This further verifies that the MANFE can perform the efficient MLE even under highly impulsive situations.

Moreover, for the low-complexity detectors in Figs. \ref{fig:bervsnr1}-\ref{fig:bervsnr3}, we can observe that the BER performance gap of the G-GAMP(30)-MANFE over the G-GAMP(30) and DetNet(30) enlarges with the imcreasing SNR. Specifically, when $\alpha=1.9$ where the noise is nearly Gaussian, the G-GAMP(30)-MANFE(1) can reduce the detection error of G-GMAP(30) to about $89.7$\%, $63.9$\% and $51.9$\% at the SNR levels of $10$ dB, $15$ dB and $20$ dB, respectively. For the same situation with regards to the DetNet(30), the above results are about $97$\%, $78$\% and $61$\%, respectively. In general, the SNR gains of the G-GAMP(30)-MANFE(1) over the G-GAMP(30) are $6$ dB, $4$ dB and $3$ dB at the BER level of $10^{-2}$ for $\alpha=1.9$, $\alpha=1.5$ and $\alpha=1.1$, respectively. In addition, the SNR gains of the G-GAMP(30)-MANFE(1) over the DetNet(30) are $9$ dB, $4$ dB and $3$ dB at the BER level of $10^{-2}$ for $\alpha=1.9$, $\alpha=1.5$ and $\alpha=1.1$, respectively. More importantly, when $\alpha=1.1$ and $\alpha=1.5$, the G-GAMP(30)-MANFE(1) has almost the same BER performance as the E-MLE in a wide range of SNR. This indicates that the G-GAMP(30)-MANFE(1) can obtain the same BER performance as the E-MLE under highly impulsive environments. In further, we can see from Fig. \ref{fig:bervsnr1}-\ref{fig:bervsnr3} that the G-GAMP(30)-MANFE(2) with a moderate computational complexity achieves the BER performance at least not worse than the E-MLE, for MIMO systems in impulsive noise environments. For example, the SNR gains of the G-GAMP(30)-MAFNet(2) over the E-MLE are about $5$ dB and $6.5$ dB at the BER level of $10^{-2}$ for $\alpha=1.5$ and $\alpha=1.1$, respectively. In particular, when the MIMO system is significantly affected by the impulsive noise, the BER performance of the G-GAMP(30)-MANFE(2) will be much better than that of the E-MLE.

\begin{figure}[t!]
    \centering
    \includegraphics[width=3in]{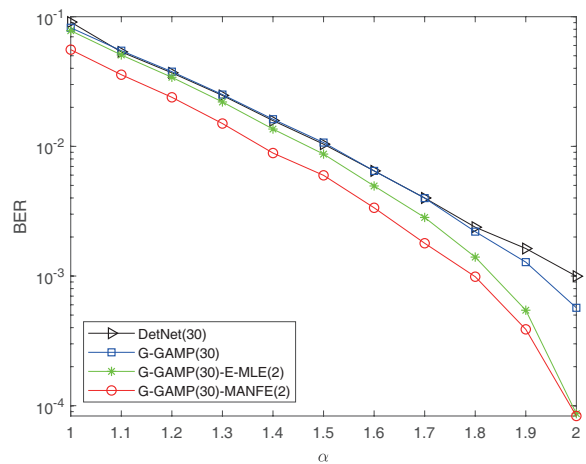}
    \caption{BER performance comparison versus $\alpha$ with SNR=25 dB for $8 \times 8$ MIMO systems.}\label{fig:bervsalpha88}
\end{figure}

\begin{figure}[t!]
    \centering
    \includegraphics[width=3in]{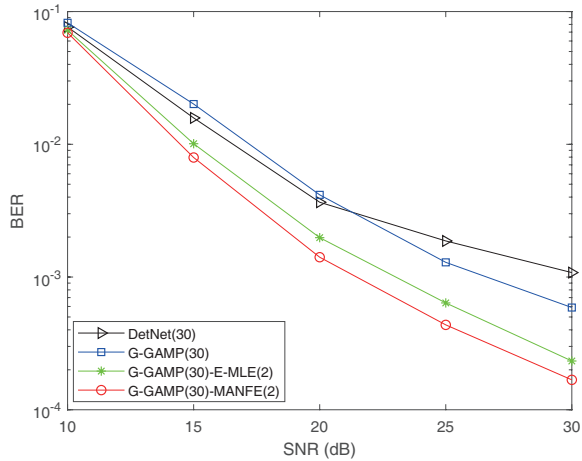}
    \caption{BER performance comparison versus SNR with $\alpha=1.9$ for $8 \times 8$ MIMO systems. }\label{fig:bervsnr881}
\end{figure}

\begin{figure}[t!]
    \centering
    \includegraphics[width=3in]{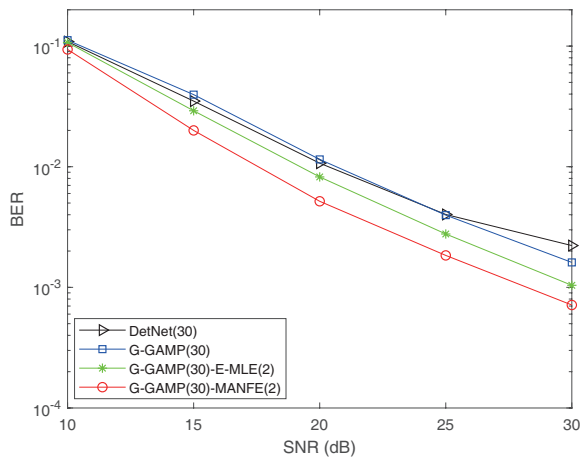}
    \caption{BER performance comparison versus SNR with $\alpha=1.7$ for $8 \times 8$ MIMO systems. }\label{fig:bervsnr882}
\end{figure}

To further verify the effectiveness of the low-complexity version of the MANFE, we use Figs. \ref{fig:bervsalpha88} - \ref{fig:bervsnr882} to demonstrate the BER performance of $8 \times 8$ MIMO systems in impulsive noise environments. Specifically, Fig. \ref{fig:bervsalpha88} shows the BER performance versus $\alpha$ with SNR $=25$ dB, while Fig. \ref{fig:bervsnr881} and Fig. \ref{fig:bervsnr882} correspond to the BER performance versus SNR with $\alpha=1.9$ and $\alpha=1.7$, respectively. In these figures, we did not plot the BER performances of E-MLE and MANFE, due to that the computational complexities of these two detectors are too high to implement in practice for $8 \times 8$ MIMO systems. Instead, we plot the BER performance of the G-GAMP-E-MLE for comparison. From Fig. \ref{fig:bervsalpha88}, we can observe that the BER performance of G-GAMP(30)-MANFE(2) is much better than that of the other detectors when $\alpha$ varies from $1$ to $2$. Specifically, when $\alpha=1.9$ and SNR$=25$ dB, the G-GAMP(30)-MANFE(2) can sufficiently reduce the detection error of the G-GAMP(30)-E-MLE(2), G-GAMP(30) and DetNet(30) to about 70\%, 30\% and 21.2\%, respectively. Moreover, the performance gain of G-GAMP(30)-E-MLE(2) over the G-GAMP(30) algorithm vanishes with the raise of the impulse strength, while the G-GAMP(30)-MANFE(2) still outperforms the G-GAMP(30) algorithm under significantly impulsive noise environments. In further, the G-GAMP(30)-MANFE(2) achieves the performance bound of the G-GAMP(30)-E-MLE(2) when the noise falls into Gaussian distribution. This further indicates that the proposed detection framework has the ability to effectively approximate the unknown distribution.

In addition, from Figs. \ref{fig:bervsnr881}-\ref{fig:bervsnr882}, we can find that the G-GAMP(30)-MANFE(2) still outperforms the other low-complexity detectors in a wide range of SNR. More specifically, when $\alpha=1.9$ where the noise is nearly Gaussian distributed and slightly impulsive, the SNR gains of the G-GAMP(30)-MANFE(2) over the G-GAMP(30)-E-MLE(2), G-GAMP(30) and DetNet(30) are $1.8$ dB, $4.5$ dB and $7$ dB at the BER level of $10^{-4}$, respectively. In the case that $\alpha=1.7$ and the noise becomes more impulsive, the SNR gains of G-GAMP(30)-MANFE(2) with respect to the G-GAMP(30)-E-MLE(2), G-GAMP(30) and DetNet(30) are $2$ dB, $4.5$ dB and $6$ dB at the BER level of $10^{-3}$, respectively. These results verify the effectiveness of the proposed framework furthermore.

\begin{figure}[t!]
    \centering
    \includegraphics[width=3in]{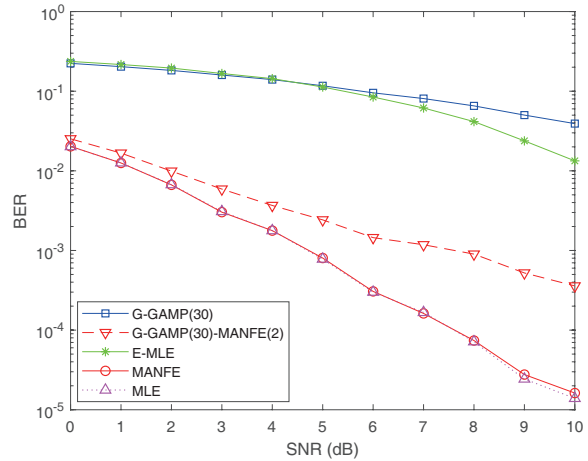}
    \caption{BER performance comparison versus SNR for $4 \times 4$ MIMO systems with Nakagami-$m$ noises. }\label{fig:bervsnr_naka}
\end{figure}

\begin{figure}[t!]
    \centering
    \includegraphics[width=3in]{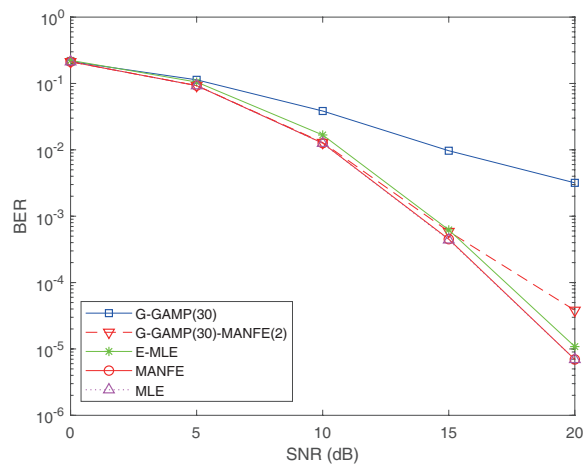}
    \caption{BER performance comparison versus SNR for $4 \times 4$ MIMO systems with Gaussian mixture noises. }\label{fig:bervsnr_mixgauss}
\end{figure}

In order to further exam the effectiveness of the proposed detection framework under different non-Gaussian noise environments, Figs. \ref{fig:bervsnr_naka}-\ref{fig:bervsnr_mixgauss} illustrate the BER performance comparisons among several detectors for the $4 \times 4$  QPSK modulated MIMO system, where the noises are subject to the Nakagami-$m$ ($m=2$) distribution and Gaussian mixture distribution Fig. \ref{fig:bervsnr_naka} and Fig. \ref{fig:bervsnr_mixgauss}, respectively. In particular, the ML estimates are available and provided in the two figures, since the two distributions are both analytical. In Fig. \ref{fig:bervsnr_naka}, we can observe that MANFE can still achieve the optimal ML performance, while E-MLE fails to work in nakagami-$m$ noise environments. As to the low-complexity detectors, G-GAMP-MANFE can still outperform the conventional detectors in nakagami-$m$ noise environment. Similar to the results in Fig. \ref{fig:bervsnr_naka}, the simulation results in Fig. \ref{fig:bervsnr_mixgauss} show that MANFE can still achieve almost the optimal ML performance under the Gaussian mixture noise environment, and G-GAMP-MANFE can still outperform the conventional sub-optimal detectors. From the simulation results of Gaussian, impulsive and nakagami-$m$ distributed noises in Figs. \ref{fig:bervsalpha}-\ref{fig:bervsnr_mixgauss}, we can find that the proposed method can almost achieve the optimal ML performance under various analytical noises, and it also outperforms the conventional detectors under non-analytical noises, which further verified the generality, flexibility, and effectiveness of the proposed method.

\section{Conclusions}\label{sec:conclusion}
In this paper, we have investigated the signal detection problem in the presence of noise whose statistics is unknown. We have devised a novel detection framework to recover the signal by approximating the unknown noise distribution with a flexible normalizing flow. The proposed detection framework does not require any statistical knowledge about the noise since it is a fully probabilistic model and driven by the unsupervised learning approach. Moreover, we have developed a low-complexity version of the proposed detection framework with the purpose of reducing the computational complexity of MLE. Since the practical MIMO systems may suffer from various additive noise with some impulsiveness of nature and unknown statistics, we believe that the proposed detection framework can effectively improve the robustness of the MIMO systems in practical scenarios. Indeed, to the best of our knowledge, the main contribution of this work is that our methods are the first attempt in the literature to address the maximum likelihood based signal detection problem without any statistical knowledge on the noise suffered in MIMO systems.

Nevertheless, there are still some interesting issues for future researches. One is that, although the low-complexity version's performance is much better than the other sub-optimal detectors, there is still a gap between it and the ML estimate. To further improve the performance of the low-complexity method, one promising approach is to leverage the automatic distribution approaching method to improve the convergence of AMP algorithms under unknown noise environments.

\bibliographystyle{IEEEtran}
\bibliography{IEEEabrv,CRN}

\end{document}